%% file: root.tex
\documentclass[10pt,conference,a4paper]{IEEEtran}

\ifCLASSOPTIONcompsoc
  \usepackage[nocompress]{cite}
\else
  \usepackage{cite}
\fi
\ifCLASSINFOpdf
\else
\fi

\usepackage{float}
\usepackage{multirow}
\usepackage{graphicx}
\usepackage{color}
\usepackage{array}
\newcolumntype{P}[1]{>{\centering\arraybackslash}p{#1}}
\usepackage{balance}

\begin{document}
%
\title{Simultaneous Food Localization and Recognition}


\author{\IEEEauthorblockN{Marc Bola\~nos and Petia Radeva}
\IEEEauthorblockA{Department of Mathematics and Informatics\\
Universitat de Barcelona,
Barcelona, Spain\\
Computer Vision Center, Bellaterra, Spain\\ 
Email: marc.bolanos@ub.edu}}


 


\maketitle

\begin{abstract}
The development of automatic nutrition diaries, which would allow to keep track objectively of everything we eat, could enable a whole new world of possibilities for people concerned about their nutrition patterns. With this purpose, in this paper we propose the first method for simultaneous food localization and recognition. Our method is based on two main steps, which consist in, first, produce a food activation map on the input image (i.e. heat map of probabilities) for generating bounding boxes proposals and, second, recognize each of the food types or food-related objects present in each bounding box. We demonstrate that our proposal, compared to the most similar problem nowadays - object localization, is able to obtain high precision and reasonable recall levels with only a few bounding boxes. Furthermore, we show that it is applicable to both conventional and egocentric images.
\end{abstract}


%
\IEEEpeerreviewmaketitle

\input{1_introduction}

\input{2_related_work}
\input{3_methodology}
\input{4_results}
\input{5_conclusion}


\ifCLASSOPTIONcompsoc
  \section*{Acknowledgments}
\else
  \section*{Acknowledgment}
\fi

Work partially funded by TIN2015-66951-C2-1-R, SGR 1219 and an ICREA Academia’2014 grant. We acknowledge NVIDIA for the donation of a GPU and M. \'Angeles Jim\'enez for her collaboration.



\bibliographystyle{plain}
\bibliography{0_main}
%
%
%

\end{document}

%% file: 1_introduction.tex
\section{Introduction}\label{sec:introduction}


The analysis of people's nutrition habits is one of the most important mechanisms for applying a thorough monitorisation of several medical conditions (e.g. diabetes, obesity, etc.) that affect a high percentage of the global population. 
In most of the cases, interventional psychologists ask people to keep a manual detailed record of the daily meals ingested. However, as proved in \cite{lichtman1992discrepancy},  usually people tend to underestimate the quantity of food intake up to a 33\%. Hence, methods for automatically logging one's meals could not only make the process easier, but also make it objective to the user's point of view and interpretability.

\begin{figure}[!t]
  \centering
      \includegraphics[width=0.8\columnwidth]{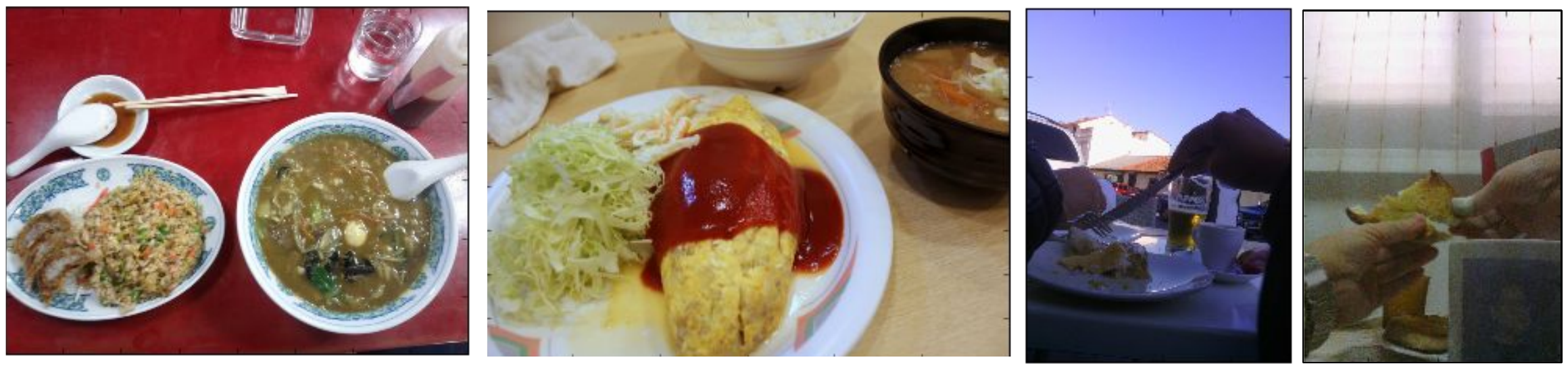}
  \caption{Examples of conventional food images (two on the left) and egocentric food images (two on the right).}
  \label{fig:samples}
\end{figure}

One of the solutions adopted recently that could ease the automatic construction of  nutrition diaries is to ask individuals to take photos with their mobile phones \cite{aizawa2013food}. An alternative technique is visual lifelogging \cite{bolanos2015towards} that consists of using a wearable camera that automatically captures pictures from the user point of view (egocentric point of view) with the aim to analyse different patterns of his/her daily life and extract highly relevant information like nutritional habits. By developing algorithms for food detection and food recognition that could be applied on mobile or lifelogging images, we can automatically infer the user's eating pattern. However, an important consideration to take into account when working with mobile or egocentric images is that they usually are of lower quality than conventional images due to the lower quality of portable hardware components. In addition, the analysis of egocentric images is harder considering that the pictures are non-intentionally taken and from a lateral point-of-view, causing motion blurriness, important partial occlusions, and bad lighting conditions (Fig. \ref{fig:samples}).

\begin{figure}[!ht]
  \centering
      \includegraphics[width=0.8\columnwidth]{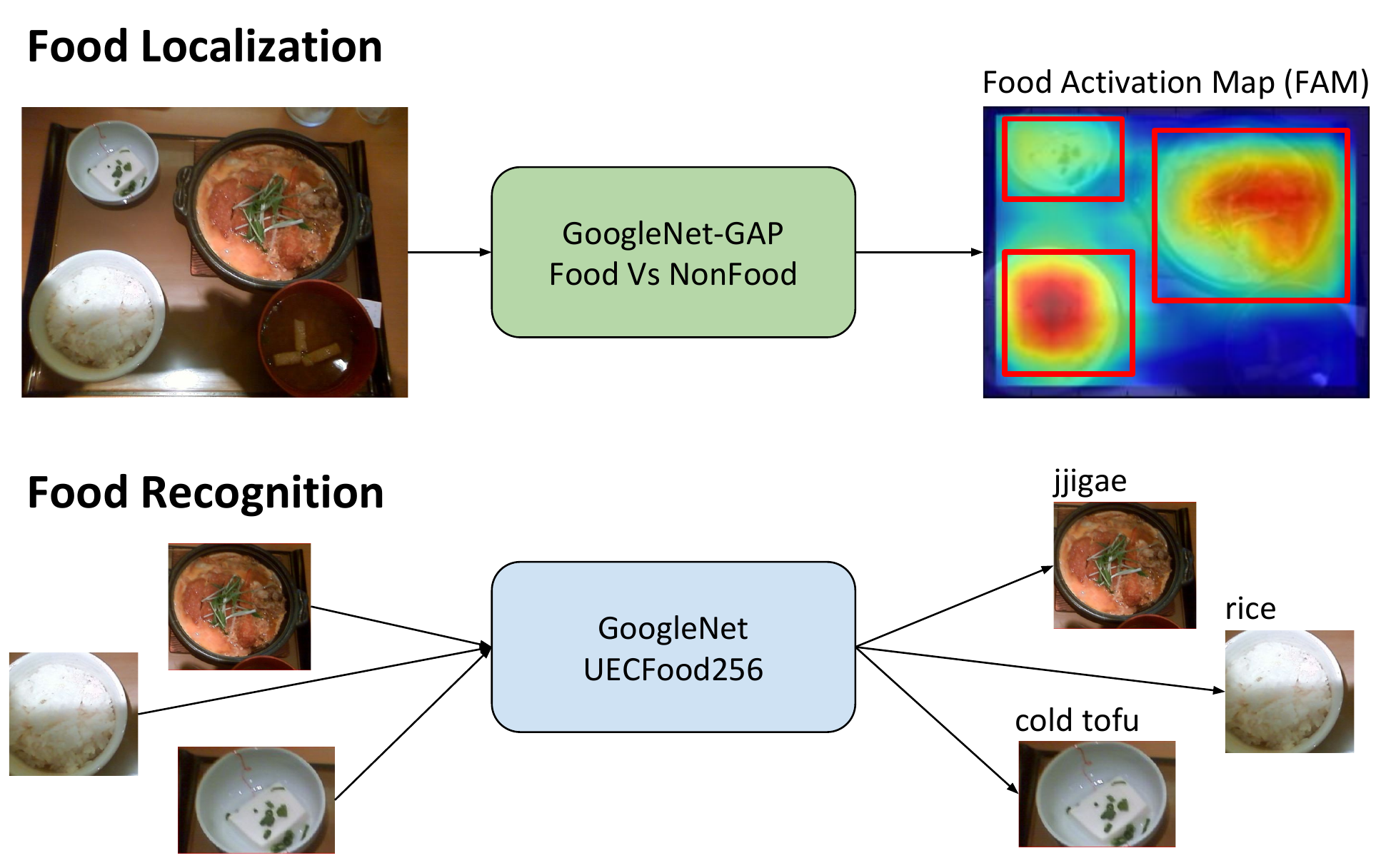}
  \caption{General scheme of our food localization and recognition proposal.}
  \label{fig:method_scheme}
  \vspace{-1em}
\end{figure}

A relatively recent technology that can leverage the automatic construction of nutrition diaries is Deep Learning, and more precisely, from the Computer Vision side, Convolutional Neural Networks (CNNs) \cite{krizhevsky2012imagenet}. These networks are able to learn complex spatial patterns from images. Thanks to the appearance of huge annotated datasets, the performance of these models has burst, allowing to improve the state of the art of many Computer Vision problems.

\begin{figure*}[!ht]
  \centering
      \includegraphics[width=0.8\textwidth]{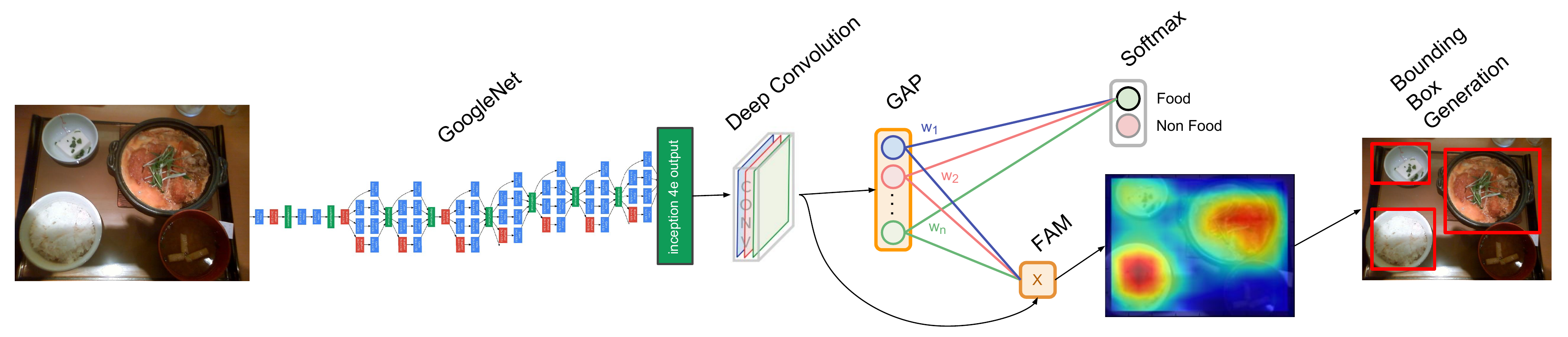}
  \caption{Our localization method based on Global Average Pooling (GAP), which produces a Food Activation Map (FAM).}
  \label{fig:food_loc}
  \vspace{-1em}
\end{figure*}

In this paper, we propose a novel and fast approach based on CNNs for detecting and recognizing food in both conventional and egocentric vision pictures. Our contributions are four-fold: 1) we propose the first food-related objects localization algorithm, which is specifically trained to distinguish images containing generic food and has the ability to propose several bounding boxes containing generic food (without particular classes) in a single image, 2) we propose a food recognition algorithm, which learns by re-using food-related knowledge and can be applied on the top of the food localization method, 3) we present the first egocentric dataset for food localization and recognition, and 4) we demonstrate that our methodology is useful for both conventional and egocentric pictures. 
Our contribution for food localization, inspired by the food detection method in \cite{kagaya2015highly}, starts by training a binary food/non food CNN classifier for food detection and then, a simple and easy to interpret mechanism that allows us to generate food probability maps \cite{zhou2015learning} is learned at the top of it. Finally, we propose an optimized method for generating bounding boxes on the obtained maps. Note that, as the desired application of the method is the generation of automatic nutrition diaries, we should not only detect food, but also food-related objects (e.g. bottles, cups, etc.). With this in mind, we collected data from complementary and varied datasets containing either food and non food pictures (see section \ref{sec:data_preprocessing}).
Up to our knowledge, there is no work in the literature that considers these categories. Without loss of generality, we add to the food categories those related to food-related objects, referring all of them as {\em food}. 
On the food recognition part, inspired by the findings in \cite{yanai2015food}, we prove that, when we have small datasets for our problem, we can apply transfer learning by performing a chain of fine-tunings on a CNN for getting closer to our target domain (food types or food-related objects recognition) and achieving a better performing network. 

The organization of this paper is as follows. In section \ref{sec:related_work}, we review the state of the art in food detection/localization and recognition. In section \ref{sec:methodology}, we explain the proposed methodology. In section \ref{sec:results}, we describe the datasets used, the experimental setup, and present and discuss our results. Finally, in section \ref{sec:conclusion}, we review the contributions, the limitations of the method and future directions.

%% file: 2_related_work.tex
\section{Related Work}\label{sec:related_work}

Considering that no works have been presented yet for simultaneous food localization and recognition in the bibliography, following we will review the most recent works devoted to food detection and food recognition, separately.


\textbf{Food Detection and Localization}: the problem of food detection has been typically addressed as a binary classification problem, where the algorithm simply has to distinguish whether a given image is representing food or not \cite{kagaya2015highly, aizawa2013food, kagaya2014food}. 
A different approach is applied by several papers \cite{anthimopoulos2013segmentation, zhu2011segmentation, meyers2015im2calories, bettadapura2015leveraging}, where they intend to first segment and separately classify the components or ingredients of food and then apply a joint dish recognition.

The main problem of both approaches is that they assume that the dish was previously localized and therefore it is centered in the image. Instead, in the context of \textit{food localization}, we are interested in finding the precise generic regions (or bounding boxes) in an image where any kind of food is present.

Although no methods have been presented specifically for food localization, several works have focused on generic object localization, usually called object detection, too. These methods could be used as a first step for food localization if they are followed by a food/non food classification applied on the obtained regions.
Selective Search \cite{uijlings2013selective}, considered as one of the best in the state of the art, applies a hierarchical segmentation and grouping strategy to find objects at different scales.
The object detection methods, which obtain generic object proposals, 
intend to detect as many objects in the image as possible for optimizing the recall level, thus, they need to propose hundreds or thousands of candidates, leading to near null precision. An open question is how to obtain straightforward object localization methods that get high precision and recall results at the same time. 
An alternative to the generic object localization methods are methods trained to localize a set of predetermined objects like Faster R-CNN \cite{ren2015faster}. The authors propose a powerful end-to-end CNN optimized for localizing a set of 20 specific object classes.


\textbf{Food Recognition}: several authors have recently focused on food recognition. Most of them \cite{ao2015adapting, ge2015modelling, bossard2014food, kawano2014foodcam, kagaya2014food, yanai2015food, kawano2013real} have analyzed which features and models are more suitable for this problem. In their works, they have tested various methods for obtaining hand-crafted features in addition to exploring the use of different CNNs. One of the best results were obtained  
in \cite{ao2015adapting} where the authors trained a CNN on the database Food101 \cite{bossard2014food} with 101 food categories and proved that applying a pre-training and then fine-tuning with in-domain food images can improve the classification performance. 
The best results on the UECFOOD256 database \cite{kawano14c} that contains 256 food categories were obtained by Yanai et al. \cite{yanai2015food}, where they used a network pre-trained on mixed food and object images for improving the final performance on food recognition. Some papers \cite{meyers2015im2calories, bettadapura2015leveraging} take a step further and use additional information like GPS location for recognizing the restaurant where the picture was taken and improve the classification results.


%% file: 3_methodology.tex
\vspace{-0.5em}
\section{Methodology}\label{sec:methodology}

In this section, we will describe the proposed methodology (see Fig. \ref{fig:method_scheme}) in two steps:  a) creating a generic food localizer, and b) training a fine-grained food recognition method by applying transfer learning.

\subsection{Generic Food Localization}

Our food-specialised algorithm detects image regions containing any kind of food, being  
reliable enough so that with a few bounding boxes it is able to keep both high precision and recall. In order to achieve a fast inference, we propose to use a CNN trained on food detection. Then, we adapt it with a Global Average Pooling (GAP) layer \cite{zhou2015learning} capable of generating Food Activation Maps (FAM) (i.e. heat maps of {\em foodness} probability). Finally, we extract candidates from the FAM in the form of bounding boxes (see pipeline in 
Fig. \ref{fig:food_loc}). 

\textbf{1) Food vs Non Food classifier}: the first step to obtain a generic food localizer is to train a CNN for binary food classification. 
We chose the GoogleNet architecture \cite{szegedy2015going} due to its proven high performance on several Computer vision tasks. We  trained the CNN on the Deep Learning framework Keras\footnote{https://github.com/MarcBS/keras}. For obtaining a faster convergence we applied a fine-tuning for our binary classification of the GoogleNet, which was previously trained on ILSVRC data \cite{ILSVRC15}.

\textbf{2) Fine-tuning for FAM generation}: once we had a model capable of distinguishing Food vs Non Food images, we applied the following steps \cite{zhou2015learning}: 1) remove the two last inception modules and the following average pooling layer from the GoogleNet for obtaining a 14x14 pixels resolution (this allows to have a high enough spatial resolution for providing a final spatial classification), 2) introduce a new deep convolutional layer with 1024 kernels of dimensions 3x3 and stride 1, 3) introduce a GAP layer that summarizes the information captured by each kernel, and 4) set a new softmax layer for our binary problem. After getting the architecture ready, we applied an additional fine-tuning for the binary problem and learning the newly introduced layers.

Note that, instead of generating a map per class as done in   Zhou et al. \cite{zhou2015learning}, we focus on obtaining a food-specific activation map that should be generic for any kind of food.

At inference time, our GoogleNet-GAP Food Vs NonFood network only has to: 1) apply a forward pass deciding whether the image contains food or not (softmax layer) and 2) compute the following equation for FAM generation:
\begin{equation}
	\textrm{FAM}(x,y) = \sum_{k} w_k \cdot f_k(x,y),
\end{equation}
where $k = \{1, ..., 1024\}$ identifies each of the kernels in the deep convolutional layer, and $w_k$ and $f_k(x,y)$ are the weighting terms of the softmax layer for the class food, and the activation of the $k$th kernel at pixel $(x,y)$, respectively.

\textbf{3) Bounding box generation}: as the last step, in order to extract bounding box proposals, we propose to apply a four steps method based on: 1) pick all regions above a certain threshold $t$, being $t$ a percentage of the maximum FAM value, 2) remove all regions covering less than a certain percentage size $s$ of the original image, 3) generate a bounding box for each of the selected regions, and 4) expand the bounding boxes by a certain percentage, $e$. All three parameters $\{t, s, e\}$ were estimated through a cross-validation procedure on the validation set (see section \ref{sec:experimental_setup}).

\subsection{Transfer Learning for Food Recognition}

After obtaining a generic object localizer, the final step in our approach is to classify each of the detected regions as a type of food. Again, for obtaining a high performing network and a faster convergence, we fine-tuned the GoogleNet pre-trained on ILSVRC. In addition, considering that our food recognition network has to overcome the problem of data quantity that most food classification datasets have, we propose applying an additional pre-training to the network. This supervised pre-training should serve as a fine-grained parameters adaptation in which the network should extract valuable knowledge from an extensive food recognition dataset before the final in-domain fine-tuning. For this purpose, we re-trained the GoogleNet, which was previously trained on ILSVRC, on the Food101 dataset \cite{bossard2014food}.

At the end, we fine-tuned the network on the target domain data (either UECFood256 \cite{kawano14c} or EgocentricFood). To obtain as little false positives as possible, we added an additional class to the final food recognition network containing Non Food samples,  enabling the system to discard false food regions detected by the localization method.

%% file: 4_results.tex
\vspace{-0.7em}
\section{Results}\label{sec:results}

In this section we will describe the different datasets used for performing the tests; the pre-processing applied to them; the metrics used for testing the localization algorithm; the experimental setup and; finally,  the results and performance of our localization and recognition techniques.

\vspace{-0.7em}
\subsection{Datasets}

Following we describe all the dataset used in this work either for food localization, for food recognition or for both.

\textbf{PASCAL VOC 2012} \cite{pascal-voc-2012}: dataset for object localization consisting of more than 10,000 images with bounding boxes of 20 different classes (none of them related to food).

\textbf{ILSVRC 2013} \cite{ILSVRC15}: dataset similar to PASCAL with more than 400,000 images and 1,000 classes for training and validation (with a subset of classes related to food).

\textbf{Food101} \cite{bossard2014food}: dataset for food recognition that consists of 101 classes of typical foods around the world, having each class 1,000 different samples.

\textbf{UECFood256} \cite{kawano14c}: dataset for food localization and recognition. It consists of 256 different international dishes with at least 100 samples each. The dataset was collected by the authors from images on the web, which means that they can be captured either by conventional cameras or by smartphones.

\textbf{Egocentric Food}\footnote{www.ub.edu/cvub/egocentricfood/}: first dataset of egocentric images for food-related objects localization and recognition. It was collected using the wearable camera Narrative Clip and consists of 9 different classes (glass, cup, jar, can, mug, bottle, dish, food, basket), totalling 5038 images and 8573 bounding boxes.

\subsection{Data Pre-processing}\label{sec:data_preprocessing}

Following we detail the different data pre-processing applied for each of the learning steps and classifiers.

\textbf{Food Vs Non Food training}: we used three different datasets: \textit{Food101}, where all the images were treated as positive samples (class Food). We used the training split provided by the authors for generating a training (80\%) and a validation (20\%) splits balanced along all classes; \textit{PASCAL}, where an object detector \cite{alexe2010object} was used to extract 50 object proposals per image on the 'trainval' set. All the resulting bounding boxes were treated as negative samples (class Non Food). Again, we divided the data in 80/20\% for training and validation; and \textit{ILSVRC}, where we selected the 70 classes (or synsets) of food or food-related objects available. In this case, we only used the training/validation split provided by the authors. The bounding boxes were extracted and used as positive samples (class Food).

\textbf{Food Recognition training}: we used the Food101 dataset as the first dataset for fine-tuning the food recognition network pre-trained on ILSVRC. The previously applied 80/20\% split of the training set provided by the authors was used for training and validation, respectively. The test set provided was used for testing. On the second fine-tuning, the same pre-processing was applied on both UECFood256 and EgocentricFood: a random 70/10/20\% split of images was applied for training/validation/testing on each class separately and the bounding boxes were extracted.

\textbf{Joint Localization and Recognition tests}: the previous 70/10/20\% split was also used on the localization and recognition test. We made sure that any image containing more than one instance was included only in one split.

\subsection{Localization Metrics}
The metric used for evaluating the results of a localization algorithm is the Intersection over Union (IoU). This metric defines how precise is the predicted bounding box (bb) with respect to the ground truth (GT) annotation, and is defined as:
\begin{equation}
	\textrm{IoU(bb)} = \frac{GT \cap bb}{GT \cup bb},
\end{equation}
where usually a bounding box is considered valid when its $\textrm{IoU} \geq 0.5$.
The other evaluation metrics used are: 
$\textrm{Precision} = \frac{TP}{TP+FP}$, $\textrm{Recall} = \frac{TP}{TP+FN}$, and $\textrm{Accuracy} = \frac{TP}{TP+FP+FN}$,
where the true positives (TP) are the bounding boxes correctly localized, the false positives (FP) are the predicted bounding boxes that do not exist in the ground truth, and the false negatives (FN) are the ground truth samples that are lost by the model. Note that given the convention from \cite{pascal-voc-2012}, if more than one bounding box overlaps the same GT object, only one will be considered as TP, the rest will be FPs.

\begin{table*}[th]
  \centering
  	\caption{Food recognition results on each dataset. Best top-1 results are shown in boldface.}
    \label{tab:food_rec}
	\begin{tabular}{ l  l  P{1.5cm} P{1.5cm}  P{1.5cm} P{1.5cm}}
 \multirow{2}{*}{\textbf{Dataset}} & \multirow{2}{*}{\textbf{Pre-training}} & \multicolumn{2}{c}{\textbf{Validation Accuracy}} & \multicolumn{2}{c}{\textbf{Test Accuracy}} \\
 & & Top-1 & Top-5 & Top-1 & Top-5 \\ \hline \hline
Food101   & ILSVRC & \textbf{74.75} & 91.11 & \textbf{79.20} & 94.11 \\ \hline
\multirow{2}{*}{UECFood256}  & ILSVRC & 52.72 & 78.61 & 51.60 & 78.26 \\
 & Food101 & \textbf{65.71} & 86.40 & \textbf{63.16} & 85.57 \\ \hline
\multirow{2}{*}{EgocentricFood} & ILSVRC & \textbf{91.50} & 99.80 & 90.77 & 99.37 \\
 & Food101 & 90.85 & 99.65 & \textbf{90.90} & 99.37
	\end{tabular}
    \vspace{-1.5em}
\end{table*}

\begin{figure}[!ht]
  \centering
      \includegraphics[width=\columnwidth]{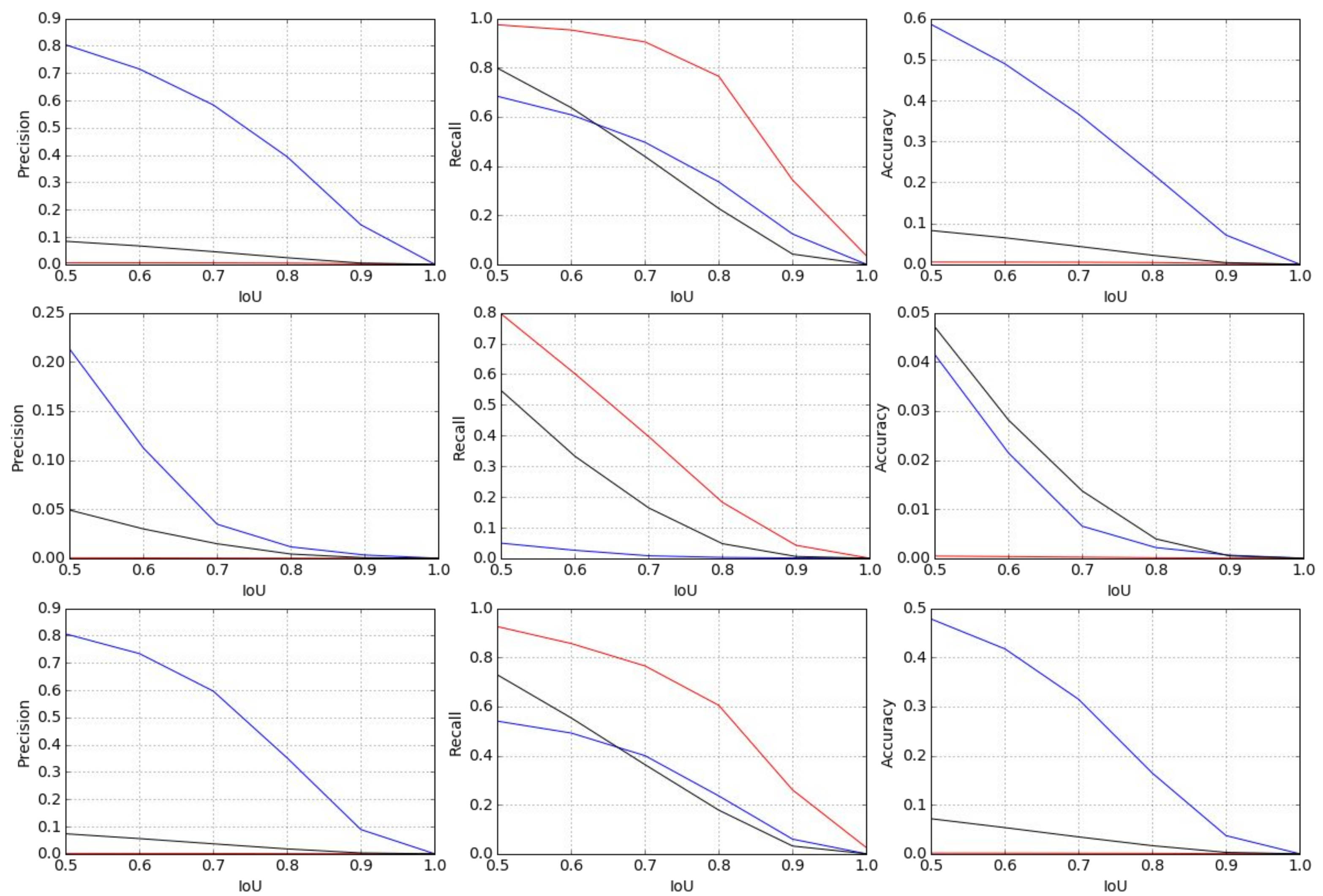}
  \caption{Curves of Precision vs IoU (left), Recall vs IoU (centre) and Accuracy vs IoU (right) on the test sets of UECFood256 (top), EgocentricFood (middle) and both combined (bottom). Our method is shown in blue, Selective Search in red and Faster R-CNN in black.}
  \label{fig:prec_rec_loc}
  \vspace{-1em}
\end{figure}

\subsection{Experimental Setup} \label{sec:experimental_setup}

The Food vs Non Food binary network used for \textit{food localization} was trained during 24,000 iterations with a batch size of 50 and a learning rate of 0.001. A decay of 0.1 was applied every 6,000 iterations. The final validation accuracy achieved on the binary problem was 95.64\%. During localization, the bounding box generation is applied on the FAM only if the image was classified as containing food by the softmax (see Fig. \ref{fig:food_loc}). 
A grid search was applied on the localization-validation set for choosing the best hyperparameters $\{t, s, e\}$ for localization (named threshold, size, and expansion percentages, respectively). The values tested were from 0.2 to 1 in increments of 0.2 for both $t$ and $e$, and from 0.0 to 0.1 in increments of 0.02 for $s$.

Considering that no food localization methods currently exist, we used Selective Search \cite{uijlings2013selective} and Faster R-CNN \cite{ren2015faster} as baselines for being two of the top performing object localization methods. The former obtains generic objects and the latter is optimized for localizing PASCAL's classes (although we will treat its predictions as generic proposals).

For the \textit{food recognition} models, first, the GoogleNet-ILSVRC model was re-trained on Food101 using Caffe \cite{jia2014caffe}, achieving the best validation accuracy after 448,000 iterations. A batch size of 16 and a learning rate of 0.001 with a decay of 0.5 every 50,000 iterations were used. The model was converted to Keras before applying the final fine-tuning to the respective datasets UECFood256 or EgocentricFood.

During the \textit{joint localization and recognition} tests, a bounding box is only considered TP if and only if it is both correctly localized (with a minimum IoU value of 0.5) and correctly recognized.

\subsection{Food Localization}

Taking into account that some of the tested methods \cite{uijlings2013selective} lack the capability of providing a localization score for each region, we are not enable to calculate a Precision-Recall curve. For this reason, we chose the accuracy as our guideline for comparison, which enables a trade-off between the capabilities of the methods to find all the objects present (Recall) and produce as little miss-localizations as possible (Precision). We chose the best $\{t,s,e\}$ parameters on the combined validation set (UECFood256 and EgocentricFood) in terms of the average accuracy value among all the IoU scores, resulting in $t=0.4$, $s=0.1$ and $e=0.2$.

In Fig. \ref{fig:prec_rec_loc} we can see the precision, recall and accuracy curves obtained by the different localization methods. 

\begin{figure}[!h]
  \vspace{-0.5em}
  \centering
      \includegraphics[trim={0 5.75cm 0 0}, clip, width=0.8\columnwidth]{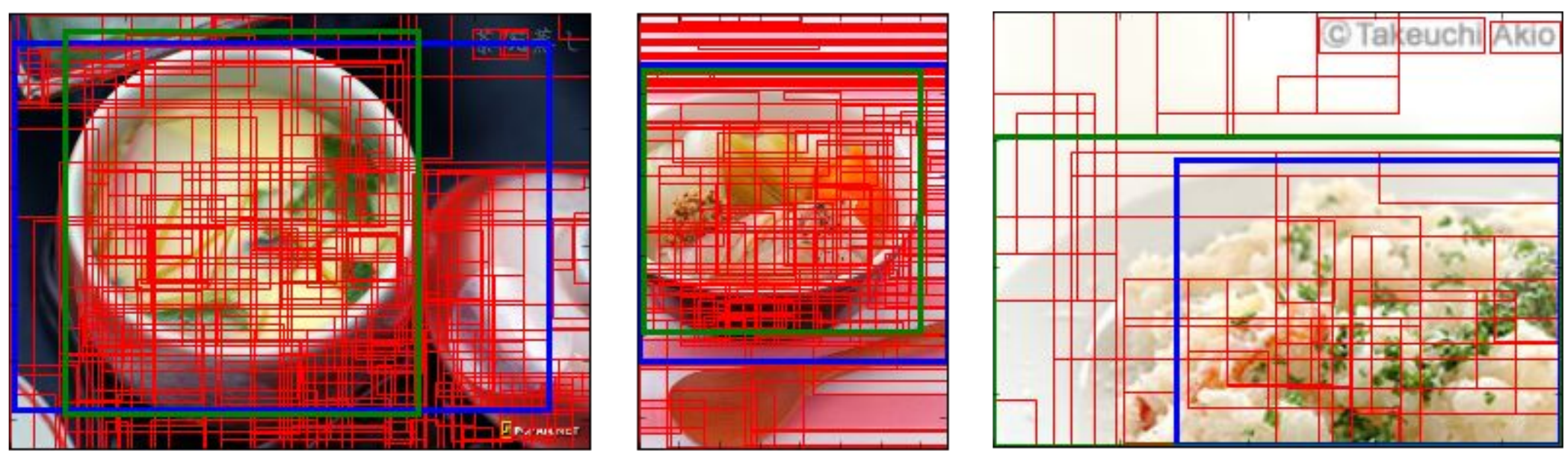}
  \caption{Examples of localization on UECFood256. Ground truth in green, our method in blue and Selective Search in red.}
  \label{fig:loc_samples}
  \vspace{-0.5em}
\end{figure}

Comparing the methods in terms of precision, it can be appreciated that ours outperforms the other methods in all cases. This pattern is easy to explain given that any generic object localization method (Selective Search in this case) usually outputs several thousands of proposals per image (see some examples in Fig. \ref{fig:loc_samples}), causing it to get a lot of FPs. In comparison, Faster R-CNN only provides some tens of proposals per image given that it is optimized for finding bounding boxes of the specific classes in the PASCAL dataset. This means that it can focus on the most interesting proposals per class, which is a great advantage compared to Selective Search and makes its precision higher. Even though, it is still far from the optimum considering that usually there are less than 10 food-related elements in an image. Note that, curiously, Faster R-CNN is able to find food-related objects even without being optimized to do so.
Comparing the methods in terms of recall, the Selective Search, in contrast to our method and Faster R-CNN, is clearly the best given that its goal is to find any object appearing in the image even if it is necessary to {\em sacrifice} the precision of the method. We can see that, although on most of the cases our method and Faster R-CNN are paired, in EgocentricFood the latter is better. This can be explained by the fact that the purpose of Faster R-CNN, which is to localize objects, is more aligned with the annotations found in EgocentricFood, which are of food-related objects.
If we compare the methods in terms of accuracy, we can see that our proposal, which is able to obtain more balanced precision-recall results, outperforms both state of the art methods in UECFood256 and the combined datasets, and is paired with Faster R-CNN on EgocentricFood.

\begin{figure*}[!ht]
  \centering
      \includegraphics[width=0.9\textwidth]{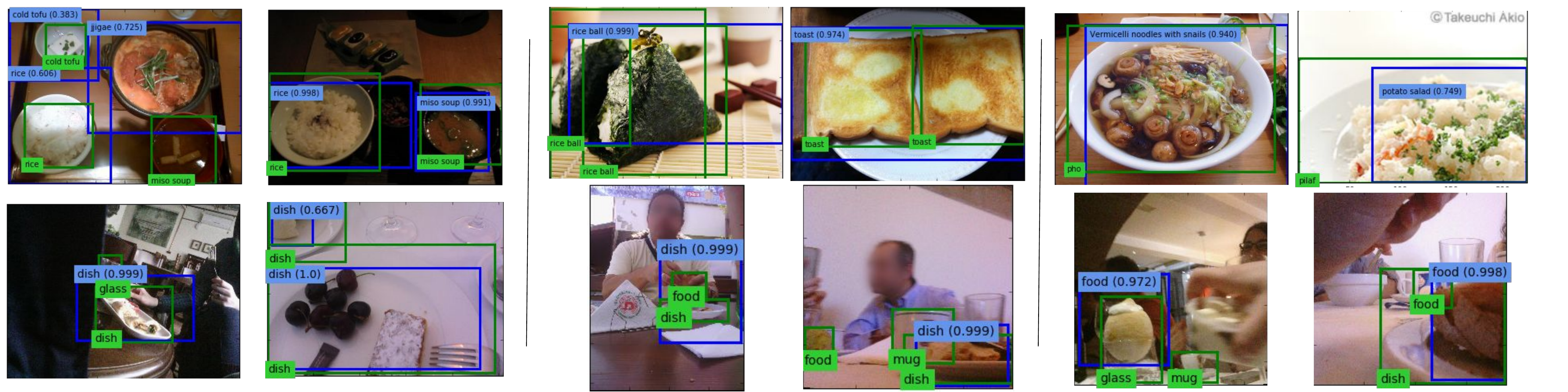}
  \caption{Examples of localization and recognition on UECFood256 (top) and EgocentricFood (bottom). Ground truth is shown in green and our method in blue (recognition score between parenthesis). Some overall good (left), good recognition, but bad localization (centre) and good localization, but bad recognition (right) examples are shown.}
  \label{fig:loc_rec_examples}
  \vspace{-1em}
\end{figure*}


As we saw, a great part of the proposed bounding boxes are correctly predicted by our method. Although, we could say that this ability is also its weak point in terms of recall, where it obtains lower values considering it is not always able to find all the food-related elements in the image, mostly when they are very close or overlapping.

Additionally, comparing them in terms of execution time, Selective Search needs an average of 0.8s per image, Faster R-CNN needs 0.2s and our localization method needs only 0.06s using a GPU and a batch size of 25. Thus, it is able to apply a near real-time inference.

\subsection{Food Recognition}

From the food recognition side, the results on the different trainings performed can be seen on Table \ref{tab:food_rec}. Note that the results are comparable to the state of the art on food recognition: either on Food101 \cite{yanai2015food}, or in UECFood256, where an alternative would be to apply the method on \cite{ao2015adapting}. We can see that, when fine-tuning on a model which is already adapted for food recognition, we can obtain better accuracy. The difference is more remarkable on UECFood256 because all the samples in the dataset are different types of food, while EgocentricFood is more focused on food-related objects.

\subsection{Localization and Recognition}

Finally, we test the whole localization and recognition pipeline proposed. We present the final results fixing the minimum IoU to $0.5$ in Table \ref{tab:loc_rec}. To take into account the results of both steps at the same time, we evaluated the precision, recall and accuracy separately for each class and applied a final mean over all the classes. Note that when combining both datasets, we have a total of 265 classes (256 on UECFood256 and 9 on EgocentricFood). Our method is able to find most of the food-related objects in the UECFood256 dataset with only a few bounding boxes (usually at most 5). On the EgocentricFood dataset the difficulty of the problem becomes clear, where there are three additional issues to overcome: 1) the quality of the pictures is lower and objects are taken in a lateral point of view, 2) some classes are ambiguous and difficult to distinguish from non food-related objects and, 3) a great part of the samples are occluded and far from the camera wearer (see examples in Fig. \ref{fig:samples} and \ref{fig:loc_rec_examples}).

Finally, in Fig. \ref{fig:loc_rec_examples} we show some examples of the complete method. In some cases, the GT ambiguity produces recognition or localization misclassification. For instance, in the first image at the bottom right zone we can see a glass (GT) with a lemon (food prediction) inside, and in the second one, we can see a dish in the foreground (GT) and a bounding box of bread in the dish (food prediction).

\begin{table}[!h]
  \vspace{-1em}
  \centering
  	\caption{Simultaneous test localization and recognition.}
    \label{tab:loc_rec}
	\begin{tabular}{ l  P{1.2cm} P{1.2cm} P{1.2cm}}
 Dataset & Precision & Recall & Accuracy \\ \hline
UECFood256    	& 54.33 & 50.86 & 36.84 \\
EgocentricFood 	& 17.38 & 8.72 & 6.41 \\
Combined		& 53.58 & 49.26 & 35.82
	\end{tabular}
    \vspace{-2em}
\end{table}

%% file: 5_conclusion.tex
\section{Conclusion}\label{sec:conclusion}

We  proposed the first methodology for simultaneous food localization and recognition. Our method is applicable to conventional and to egocentric point-of-view images. We have proven that this methodology outperforms the baseline achieved by generic object localizers. As future work, we will focus on the ability of the method to distinguish very close or overlapping food-related objects. 

\balance